\documentclass[letterpaper, 10 pt, conference]{ieeeconf}  

\IEEEoverridecommandlockouts                              

\usepackage[backend=bibtex,
            hyperref=false,
            url=false,
            isbn=false,
            doi=false,
            backref=false,
            style=numeric-comp,
            sortcites,
            block=none]{biblatex}

\usepackage[english]{babel}
\usepackage[T1]{fontenc}

\usepackage[nolist]{acronym}
\usepackage{amsmath}
\usepackage[font={small}]{caption}
\usepackage{subcaption}
\usepackage[export]{adjustbox}
\usepackage{amssymb}
\usepackage{bm}
\usepackage{booktabs}
\usepackage{booktabs}
\usepackage{balance}
\usepackage{csquotes}
\usepackage{color}
\usepackage{colortbl}
\usepackage{dsfont}
\usepackage{esvect}
\usepackage{float}
\usepackage{graphicx}
\usepackage{hyperref}
\usepackage{import}
\usepackage{mathtools}
\usepackage{multirow}
\usepackage{multicol}
\usepackage{multirow}
\usepackage{flushend}
\usepackage{outline}
\usepackage{paralist}
\usepackage{pifont}
\usepackage{siunitx}
\usepackage{stmaryrd}
\usepackage{systeme}
\usepackage{soul}
\usepackage{optidef}
\usepackage{url}
\usepackage{tikz}
\usepackage{tabularx}
\usepackage{verbatim}
\usetikzlibrary{tikzmark}

\newcommand{\cmark}{\color{green}\ding{51}}%
\newcommand{\xmark}{\color{red}\ding{55}}%

\usepackage[normalem]{ulem}

\title{\LARGE \bf

Everyday Finger: A Robotic Finger that Meets the Needs of Everyday Interactive Manipulation

\thanks{\scriptsize{\textbf{Extended/updated version of article to appear in IEEE ICRA 2024 proceedings.}
© 2024 IEEE.  Personal use of this material is permitted.  Permission from IEEE must be obtained for all other uses, in any current or future media, including reprinting/republishing this material for advertising or promotional purposes, creating new collective works, for resale or redistribution to servers or lists, or reuse of any copyrighted component of this work in other works.}}
}

\author{
\authorblockA{Rubén Castro Ornelas, Tomás Cantú, Isabel Sperandio, Alexander H. Slocum, and Pulkit Agrawal \\ 
Improbable AI Lab $\quad$
Massachusetts Institute of Technology}
}

\begin{document}

\maketitle
\thispagestyle{empty}
\pagestyle{empty}

\begin{abstract}
We provide the mechanical and dynamical requirements for a robotic finger capable of performing thirty diverse everyday tasks. To match these requirements, we present a finger design based on series-elastic actuation that we call the \textit{everyday finger}. 
Our focus is to make the fingers as compact as possible while achieving the desired performance.  We evaluated \textit{everyday fingers} by constructing a two-finger robotic hand that was tested on various performance parameters and tasks like picking and placing dishes in a rack, picking thin and flat objects like paper and delicate objects such as strawberries. Videos are available at the project website: \url{https://sites.google.com/view/everydayfinger}.

\end{abstract}

\section{INTRODUCTION}
Enabling robots to perform dexterous manipulation tasks akin to the human hand has been a long standing challenge in robotics. When attempting fine, dexterous, and forceful manipulation in human environments designed around human bodies, incorporating some properties of human hands, like their size, morphology, and compliance, into a robotic manipulator can be advantageous. Building human-like multi-finger hands is well researched~\cite{bicchi2000hands,piazza2019century}, but there is lack of reliable and performant dexterous robotic hands that are available off-the-shelf. The primary challenges have been: (i) The difficulty of controlling a system with a large number of degrees of freedom and contact. However, recently, reinforcement learning has emerged as a promising method for hands~\cite{andrychowicz2020learning,handa2022dextreme,zhu2019dexterous,khandate2022feasibility,chen2022visual}. 
(ii) Properly balancing the tradeoff between the manipulator's size, compliance, durability, and the power it can provide. As an example, for the fingers to exert a large force, high torque is necessary but such actuators usually have a big form factor making the fingers bulky and thereby making the hand incapable of using many human tools purely from a geometric standpoint.

To achieve the requisite performance objectives within a compact hand, many dexterous robotic hand designs such as the Robonaut II \cite{bridgwater2012robonaut}, Shadow hand \cite{ShadowDextereous}, or DLR hand \cite{GuizzoDLR} and others~\cite{SaLoutos_Kim_Stanger-Jones_Guo_Kim_2022,Bridgwater_Ihrke_Diftler_Abdallah_Radford_Rogers_Yayathi_Askew_Linn_2012,Jacobsen_Iversen_Knutti_Johnson_Biggers_1986} house motors in a forearm and route tendons to the fingers. While this design sidesteps the challenge of miniaturizing the actuator, it adds complexity through intricate transmission systems, a cumbersome forearm, and a high manufacturing cost. Additionally, tendons must pass through a moving wrist. All these parts decrease reliability. Shadow dexterous hand, one of the most popular robotic hands, is known to be a challenge to work with due to breakages and maintenance being  complicated~\cite{openai2019learning} due to the complex design. For these reasons, we avoided the use of a tendon-driven mechanism. 

\begin{figure}[t!]
    \centering
    \includegraphics[width=\linewidth]{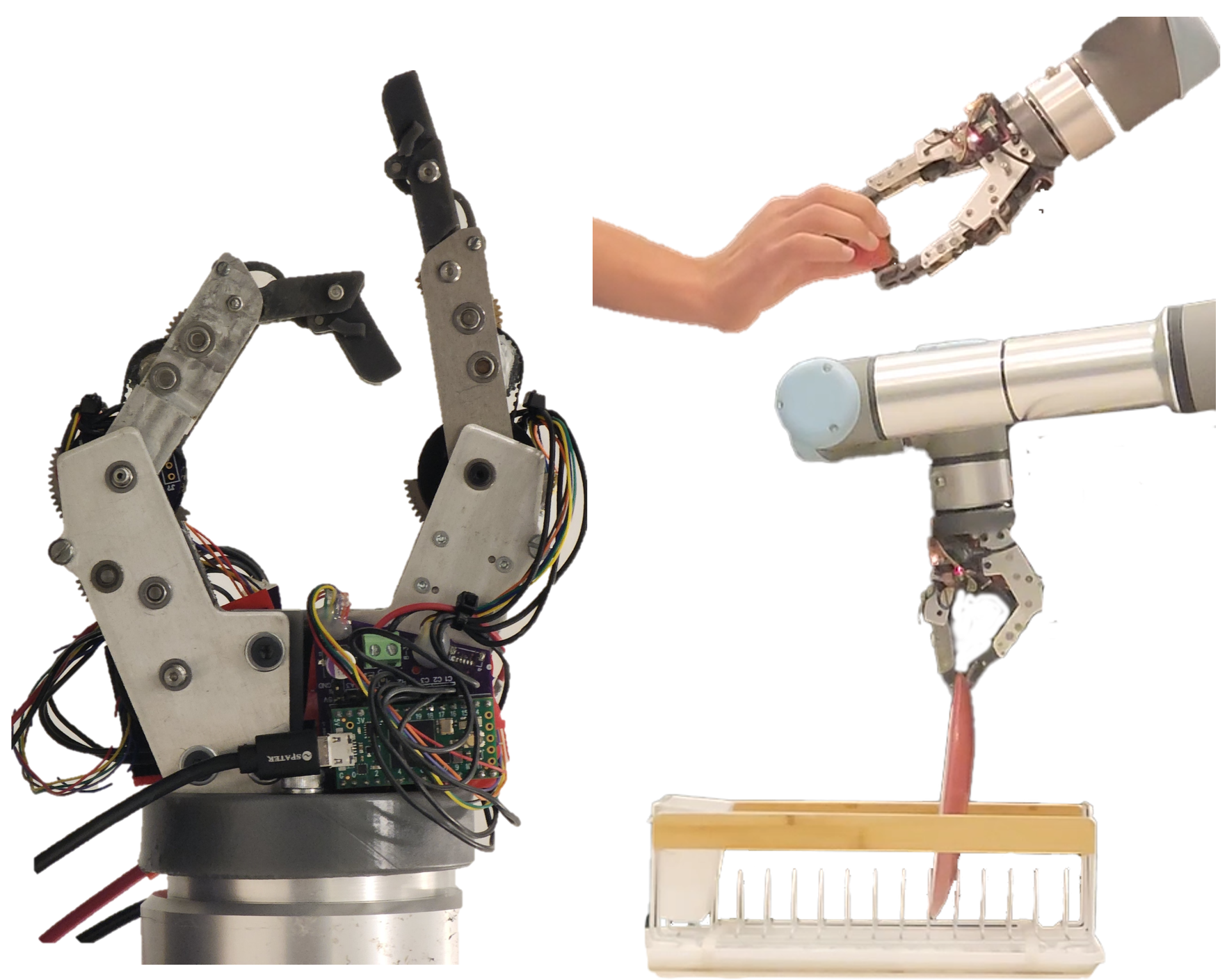}
    \caption{The proposed design of the \textit{everyday} two-finger robotic hand that is appropriate for daily tasks.}
    \label{fig:our_hand}
\end{figure}%

A range of recent robotic hands, like the Allegro hand~\cite{Allegro_Hand} and the Leap Hand~\cite{shaw2023leap}, use off-the-shelf actuators to directly drive joints, emphasizing minimal custom parts to keep costs low and increase the accessibility of robotic hands. While these models serve research objectives effectively, their reliance on off-the-shelf components often results in a bulkier build (1.5-3x times the human hand), which complicates their utility for manipulating tools designed for human use. 

The key missing piece to having a robotic hand that is similarly sized to a human hand is an actuator design that balances the performance trade-offs.  Qualitatively, for a robotic hand to be useful for everyday tasks, we need it to be forceful, speedy, high bandwidth, compliant, and compact. A useful robotic hand can exert enough \textbf{force} so that it can firmly hold heavy objects like a milk jug and use tools. At the same time, it needs to have enough \textbf{speed} and \textbf{bandwidth} in order for the hand to be used for dynamic tasks such as re-orientation. While one might assume it's acceptable for robotic hands to move slower than humans, finger speed serves a broader purpose. If an object slips and begins to fall during re-orientation, the fingers must swiftly adjust to catch it. Hence, quick finger movements are not just about efficiency, but also ensuring task reliability and a higher success rate in handling disturbances.  \textbf{Compactness} allows the robotic hand to use human tools. If the hand is too big, it will struggle to pick up smaller and flat objects and is impossible for it to operate certain tools such as scissors. Lastly, a useful robotic hand also has \textbf{compliance}, which enables contact-rich tasks and makes high-level planning easier. Consider reaching for a wooden block on a table: even with misalignment, compliant fingers can adapt to the table's surface, allowing it to guide the fingers to the block. This not only improves grasp reliability but also protects against potential damage from unforeseen impacts, facilitating safer and more versatile interactions with various objects and surfaces in dynamic environments. 

Our first contribution is to define and set quantitative targets for these five critical metrics, \textit{compactness, compliance, force, speed, and bandwidth}, that together assess the quality of a robotic hand for performing everyday tasks (Section~\ref{sec:functional-requirements}). Based on this analysis, our second contribution is a series elastic actuator design that matches these requirements, except for two, it is wider and taller at the knuckle than a human hand by $\sim$50\%. This may prevent it from being used in a five finger hand configuration. The reasons we can now be more compact than previous designs are: (i) our design choice prioritizing common daily tasks over extreme human capabilities (ii) the increased availability and miniaturization of torque-dense brushless electric motors. It uses off the shelf motors, but with a custom structure and transmission that uses a spring in series to lower the impact loads. Using this actuator, we built a two-finger robot hand to empirically verify its performance on key functional requirements and demonstrate its potential in performing everyday tasks (see Fig ~\ref{fig:our_hand}).

\section{Other Prior Works}
\label{sec:related-work}

\subsection{Soft hands} Soft robotic hands offer delicate handling that rigid hands might struggle with \cite{shintake2018soft}. Their inherent compliance can make them adept at tasks like handling fruit. However, accurately simulating the complex dynamics of soft materials presents a significant challenge. Traditional simulation environments may not accurately capture the nuanced behaviors of soft robots. Because of this, using the latest reinforcement learning control and Sim2Real techniques may not be possible for soft robots until simulators become faster.

\subsection{Similarities between manipulation and locomotion}
Robot quadrupeds is a field that has benefited from advancements in brushless motor technology/availability and from introducing compliance. Proprioceptive actuators, like in the MIT Cheetah \cite{wensing2017proprioceptive}, have enabled contact-rich and dynamic movements. Series Elastic Actuators (SEAs) like those in the ANYmal robot dog \cite{hutter2016anymal} have also enabled stable contact interactions with the environment, albeit with lower bandwidth. We employ a similar analytical framework used for the MIT Mini Cheetah design \cite{katz2019mini} in Section~\ref{sec:proposed-design} to design our actuator.

\section{Functional Requirements}
\label{sec:functional-requirements}
Our first goal is to characterize the strength, speed, bandwidth, and compliance requirements for a robotic hand that is as compact as a human hand and can still apply forces large enough to perform everyday tasks. We will now detail how we obtained such a characterization. The results are summarized in Table~\ref{tab:desiderata}.

\subsection{Force, Bandwidth, and Speed Characterization}
\begin{figure*}[ht]\centering
\vspace{0.2cm}
\includegraphics[width=\linewidth]{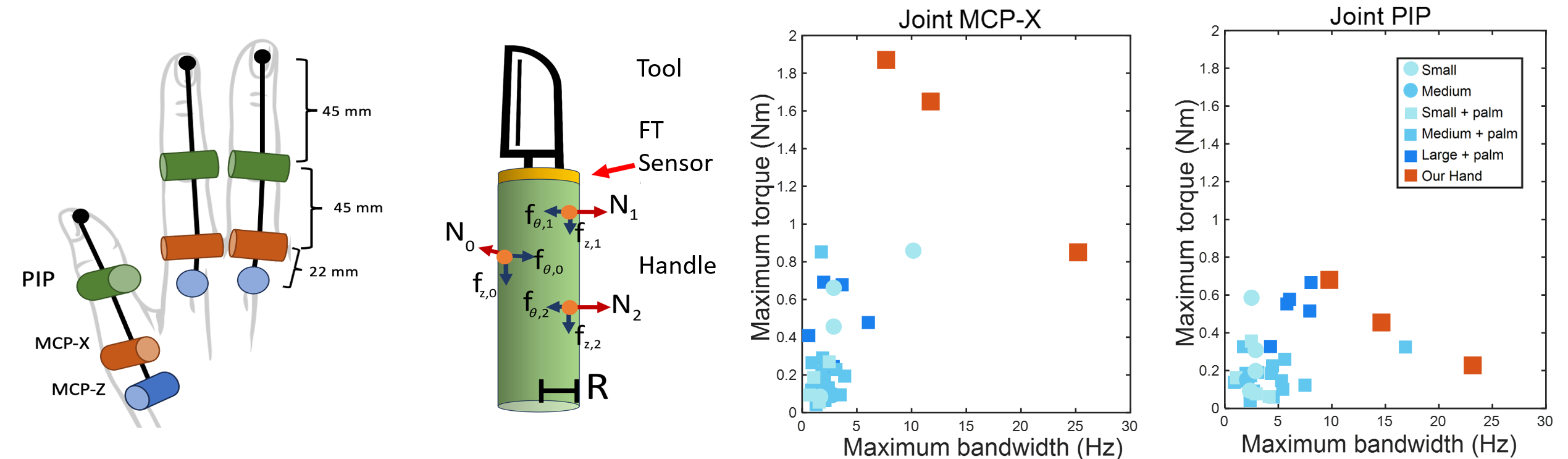}
    \caption{Left: Finger geometry and degrees of freedom used in optimization. Left-Middle: diagram of three contact points (orange) on a cylindrical tool handle and their corresponding forces as used in optimization. Right: Our hand's capabilities compared to daily task requirements. Tasks are split between small, medium and large handle radius, as well as whether 4th contact paint, a palm, was used. }
    \label{fig:Optimization}
\end{figure*}

It is important for a finger to not only apply forces and move fast but also quickly transition between different forces (i.e., bandwidth). To obtain force and bandwidth characterization for everyday tasks, we leveraged a prior study that attached a sensor to the base of tools to measure force and torque applied by humans while using the tools to perform 30 different daily tasks \cite{huang2018dataset}. These measurements do not represent the minimum effort to complete the tasks but the average effort of the person using the tool. Making a reasonable set of assumptions that we explain below, we mapped these measurements to torques experienced at each joint of a model hand with three fingers, with 3 DOFs each. The three joints per finger can be seen in Figure ~\ref{fig:Optimization}, and are the Proximal Interphalangeal (PIP), Metacarpophalangeal Z-axis (MCP-Z), and the Metacarpophalangeal X-axis (MCP-X). 

Inferring torques exerted at each joint of the \textit{model} hand based on Force-Torque (FT) sensor readings from a single sensor at the tool base is an under-specified problem with many possible solutions. Consequently, to solve for joint-level torques, we add multiple constraints: 

(Constraint/Assumption 1) Minimize the peak torques required from motors to be able to choose smaller motors: 
\begin{align}
 \text{min} \quad \sum_{i=0, 1, 2} \tau_{MCP-Z, i}^4+\tau_{MCP-X, i}^4+\tau_{PIP, i}^4 
 \label{eqn:Torque_min}
\end{align}
where $\tau_{MCPZ, i}, \tau_{MCPX, i}, \tau_{PIP, i}$ are the torques on each joint for each finger i. We found that minimizing the fourth power of the torques resulted in superior minimization of peak torques, whereas lower powers optimized average torques, but not the peak torques as well.

\begin{figure}[]
\centering
\vspace{0.1cm}
\includegraphics[width=8cm]{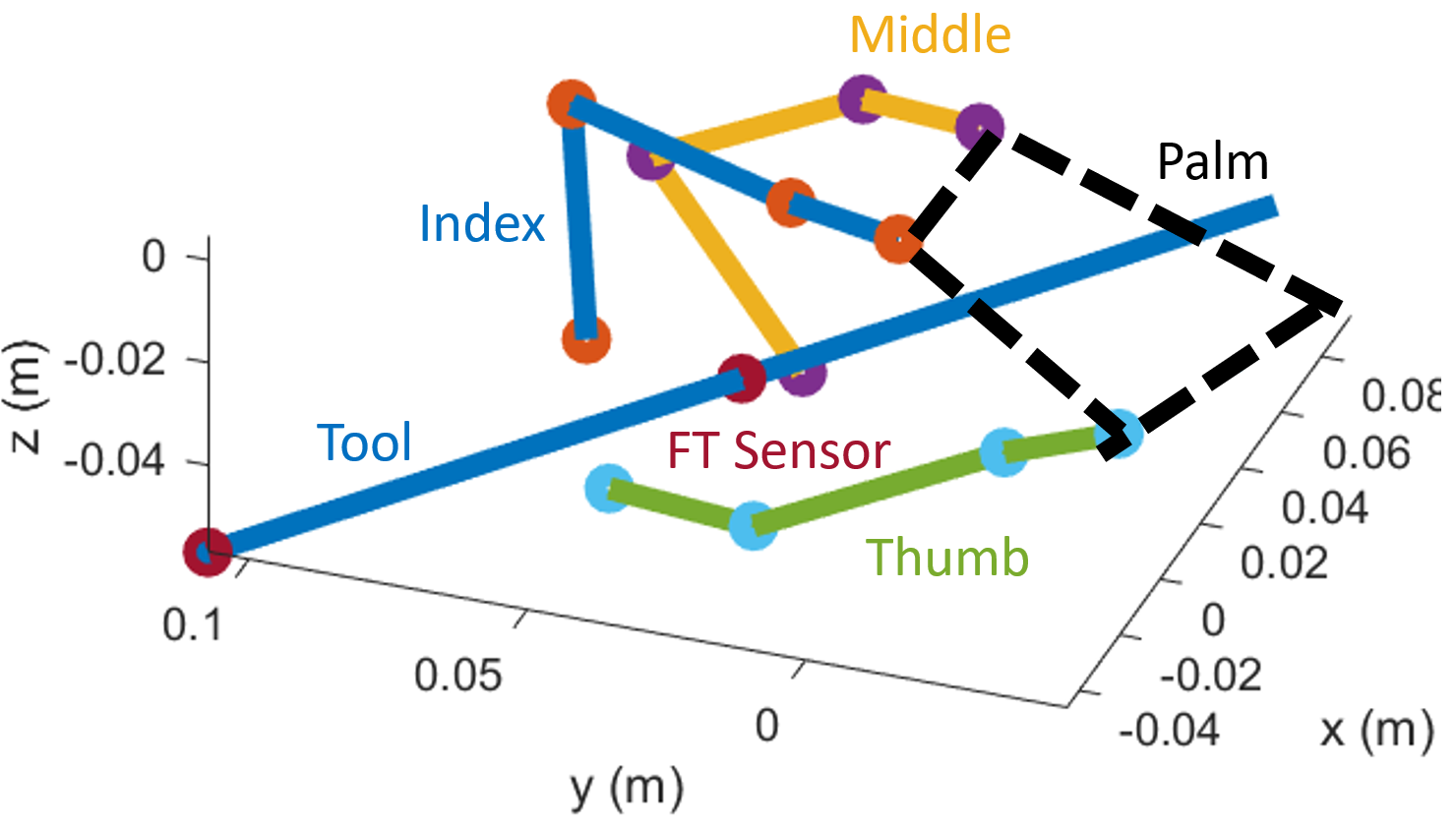}
\caption{Medium Pinch grasp, such as is used for the screwdriver tasks, for inferring torque trajectories at the joints based on FT sensor readings (shown in red in the diagram).}
\label{fig:grasp_example}
\end{figure}

(Constraint/Assumption 2) Estimating the joint torques required by the \textit{model} hand to exert the same FT reading as recorded at the tool base requires knowledge of the finger configuration and their contact location with the tool. We assume that the three fingers of the \textit{model} hand statically grip a cylindrical handle for each task. The other challenge is that the dataset used a cylinder of fixed radius as the tool handle when humans performed daily tasks. However, in the real world, humans don't use a tool handle but directly grip tools of varying shapes and sizes. To account for this variation, we performed joint-level torque calculations for tool handles of three different radii: 8mm(small), 15mm(medium), and 22mm(large). For each task, we chose the cylinder size that resembles our intuitive estimate of the size of the real-world tool (see Fig. \ref{fig:Optimization}). Beyond the three fingers, for a stronger grip, most tasks (i.e., using a hammer to hammer a nail) also incorporate a fourth contact point on the cylinder (a "palm"). The appendix details the choices of cylinder radius, whether we assumed contact with the palm, and grasp points/palm points for each task. One example is given in Fig. \ref{fig:grasp_example} These were chosen qualitatively based on how we would pick them up naturally ourselves. Once the contact configuration is fully specified, we can express the static equilibrium condition of a stable grasp with six equations, one for each axis in force and moment balance:

\begin{align}
    & F_x = \sum_{i=0, 1, 2, 3}(N_i\cos\theta_i-f_{\theta,i}\sin{\theta_i}) \label{eqn:F_x}  \\
    & F_y = \sum_{i=0, 1, 2, 3}(N_i\sin\theta_i+f_{\theta,i}\cos{\theta_i})  \\
    & F_z = f_{z, 0}+f_{z, 1}+f_{z, 2}+f_{z, 3}  \\
    & \begin{aligned} T_x &=\sum_{i=0, 1, 2, 3}(-z_i(N_i\sin(\theta_i)+f_{\theta,i}\cos{\theta_i}) \\
    &\quad +R\sin\theta_if_{z, i}) \end{aligned}  \\
    & \begin{aligned} T_y &= \sum_{i=0, 1, 2, 3}(z_i(N_i\cos\theta_i-f_{\theta,i}\sin{\theta_i}) \\
    &\quad -R\cos\theta_if_{z, i}) \end{aligned}  \\
    & T_z = R(f_{\theta, 0}+f_{\theta, 1}+f_{\theta, 2}+f_{\theta, 3}) 
\end{align}
where $F$ and $T$ are the measured forces and torques from the FT sensor, $N_i$ are the touch-point forces normal to the cylinder, the $f$ are the friction forces in $\theta$ and $z$ components of the cylinder frame, $R$ is the radius of the cylinder. $z_i, \theta_i$ are the locations of the actual touch-point forces at each time step around the cylinder. 

We conducted a sensitivity analysis to assess the impact of the precise locations of touch-points. We selected five tasks that use varying configuration of grips and cylinder radii from our database: 2-Spatula, 3-Shaker, 11-Using Fork, 13-Screwing, 23-Peeling, and randomly generated new touch-point origins within a 5mm radius of their original locations, along with changing its cylinder radius by $\pm$ 5mm. We observed that the peak torques varied by an average of 0.024$\pm$0.023 Nm, a small difference. 


(Constraint/Assumption 3) We implement a friction cone on each touch-point with a coefficient of friction of 0.6. In the real world, this varies wildly depending on skin moisture, materials being handled, age, and more. For our purposes, we chose 0.6 to be a reasonable average scenario. For reference, dry skin on glass has coefficient of friction of 2.18 $\pm$ 1.09 \cite{derler2012tribology}. Equation \ref{eqn:friction_cone} can be seen below, where $\mu$ is the assumed coefficient of friction.

\begin{align}
\label{eqn:friction_cone}
    & f_{\theta,i}^2 + f_{z,i}^2 \leq (\mu N_i)^2  \;\;\;\;\;\;\;\;\; \text{for } i=0, 1, 2, 3
\end{align}

We took the same 5 tasks from the sensitivity analysis on grasps, and ran a new sensitivity analysis changing the coefficient of friction to 0.5 or 0.7 for each. This varied the peak torques by only 0.042$\pm$0.032 Nm, showing that small perturbations in friction coefficient do not cause large differences in outcome.

(Constraint/Assumption 4) Because we apply forces across our whole fingertip and not at an individual point, We assume that the fingertip force will lie somewhere in a 8mm radius circle of the center of the fingertip, the same size as a human fingertip. This makes sense as a "center of pressure" since the contact is not a single point. The palm force point, if present, has a 12mm radius available. The implementation can be seen below, where  $\Bar{z_{i}}$, $\bar{\theta_{i}}$ are the locations of the center of the touch-point pressure areas (Subscript 3 refers to the palm, when available), and $r_{presssure}$ is the radius of the touch-point pressure area.

\begin{align}
    &\begin{aligned} (z_i - \Bar{z_{i}})^2+R^2(\theta_i - \bar{\theta_{i}})&^2 \leq r_{pressure}^2 \\
    &\quad \text{for } i=0, 1, 2, 3\end{aligned}
\end{align}

(Constraint/Assumption 5) The lengths of the links of the finger are given in Fig. \ref{fig:Optimization}, these correspond approximately to  human fingers but also to what is realistic using robot parts. In particular, we note that MCP-X and MCP-Z are usually co-located at the same point. However, to avoid using ball joints to keep things simple, we separate them by 22mm, enough space to put these joints in series, as shown in Fig. \ref{fig:Optimization}. This means that the MCP-Z joint in our design will have higher torques than those experienced by humans since the joint is further away from where the force is applied. To match the forces at the fingertips to joint torques, we use the jacobian matrix ($J$) to transfer between the fingertip frame and joint-space.

\begin{align}
    J^T &\begin{bmatrix}
        N_i \\
        f_{z, i}\\
        f_{\theta_i}\\
    \end{bmatrix}
    =
    \begin{bmatrix}
        \tau_{MCP-Z, i} \\
        \tau_{MCP-X, i}\\
        \tau_{PIP, i} \\
    \end{bmatrix} \text{for } i=0, 1, 2
    \label{eqn:jacobianeqn}
\end{align}

(Constraint/Assumption 6) The force due to gravity of the tools is measured by the FT sensor used in the force studies, but we ignore gravity effects of the fingers.  The weight of the finger contributes less than 5\% to the maximum output torque of each motor in the worst case. This is when the finger is fully stretched parallel to the ground, at which point it takes 0.02Nm to keep it up.

\begin{figure}[h]
\centering

\vspace{0.3cm}
\includegraphics[width=6cm]{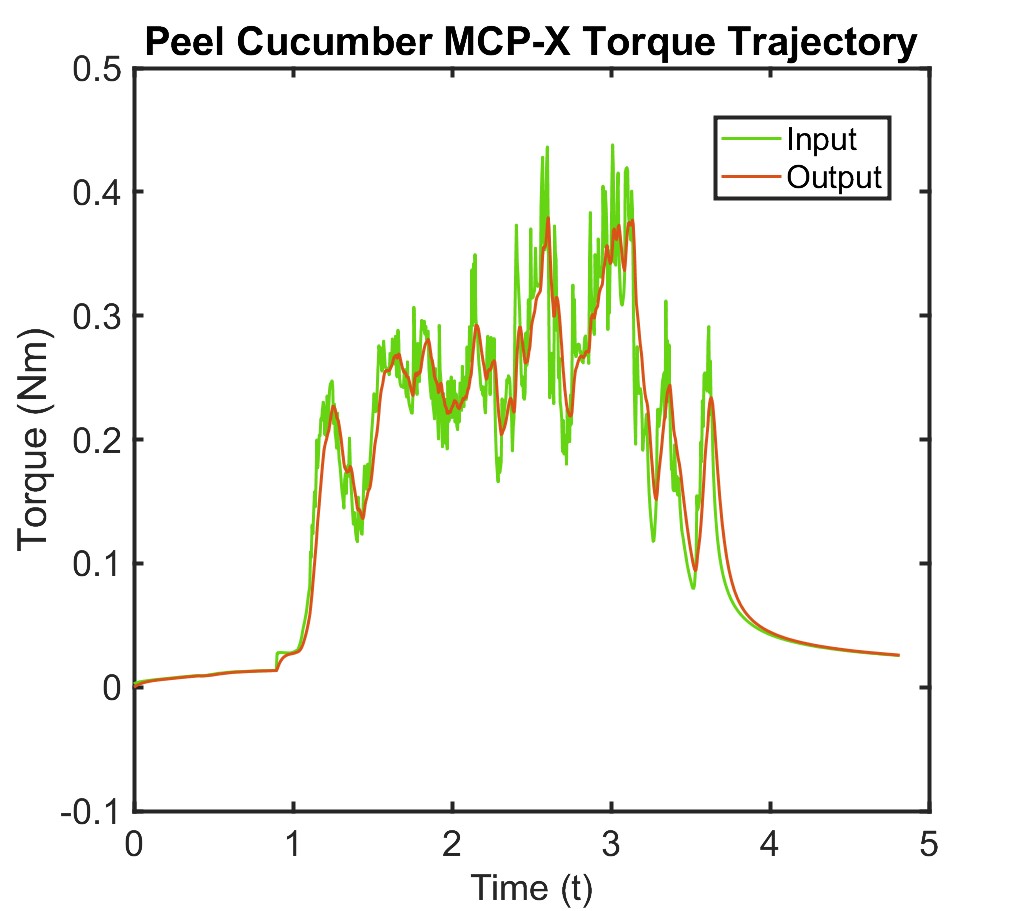}
\caption{ Optimized peeling cucumber joint torque trajectory for index finger MCP-X joint and the response of a first order system with 5Hz bandwidth, the requirement for this task based on our analysis. Note that the high frequency noise from the raw data is attenuated, but the main signal is followed closely.}
\label{fig:5Hzbandwidth}
\end{figure}

We optimized the joint torque trajectories for each of 30 tasks in the daily interactive manipulation dataset, subject to the above constraints 
One such joint torque trajectory can be seen as the input in Fig. \ref{fig:5Hzbandwidth}. We computed the maximum torque and bandwidth required to follow each trajectory closely on each of the three joint types in the analysis (PIP, MCP-Z, MCP-X). To find the requisite bandwidth, we simulate open-loop first order systems with bandwidth $B_{rad}$ in $\sim$0.2Hz increments between 0 and 100Hz, the expected range based on other robot systems. The transfer function for such system is: 
\begin{equation}
    T(s) = \dfrac{\sqrt{B_{rad}^2 +1}}{{s+B_{rad}}}
\end{equation}
To simulate how these systems perform, we use Matlab's lsim function, which can take in a system's transfer function and an input trajectory and provide the output of that system. Then we pick the system with the lowest bandwidth in which the output of the first order system given the reference signal has 98\% of the time steps within 5\% of the reference signal. We chose these parameters qualitatively after a few iterations until they attenuated noise, while closely following the signal. Results from this analysis done on a sample torque trajectory can be seen in Figure \ref{fig:5Hzbandwidth}. A summary of torque and bandwidth results from each task in the database are in Figure~\ref{fig:Optimization}

\vspace{5pt}
\noindent \textit{Speed:}
To obtain a measure of the maximum speed that a human finger can operate at, we obtained the data from a prior work analyzing the maximum speed of an expert pianist's finger while playing piano.  \cite{Castro_Ornelas_2022}. This becomes an upper bound to the speed required for everyday tasks.


\subsection{Compactness}
To ensure compatibility with human tools, our design's size constraints are grounded in a study that examined the bone and soft tissue measurements of human fingers \cite{Sun_2021_fingersize_prosthetics}. The maximum expected length of a human index finger, calculated as the mean length plus three standard deviations, is 104.2 mm for bone and an additional 5.61 mm accounting for soft tissue. For finger width, we followed the dimensions provided in \cite{Sun_2021_fingersize_prosthetics}, which align closely with typical human hand widths, giving us a width allowance of 22.4 mm per finger, including soft tissue. We use the same value of 22.4mm for the height of the finger for symmetry.

\subsection{Compliance}
Having a low torque to back-drive the actuator allows the finger to be compliant. This backdrivability  defines how much force an object being interacted with must exert on the robot fingers to push them back. Ideally, this force to push back the robot is light enough that delicate objects such as fruit can push back the fingers without damaging the fruit, or such that a finger hitting a table doesn't damage it and instead can be used as a guide. In order to handle these scenarios, we wanted the maximum force to backdrive to be 1N at the fingertip, which is light enough to not damage most surfaces. This leads to 0.1Nm at the MCP-X joint. We chose the same value for the PIP joint.

\begin{table}[t!]
    \vspace{0.35cm}
    \centering
    \caption{Comparison of Desired and Achieved Functional Requirements}
    \label{tab:requirements}
    \begin{tabularx}{\columnwidth}{X|c|c|c}
        \textbf{Functional Requirement} & \textbf{Desired} & \textbf{Ours} & \textbf{Pass} \\
        \hline
        PIP Torque (Nm) & 0.65 & 0.68$\pm$0.01  & \cmark \\
        PIP Bandwidth(Hz)@0.55Nm& 8.69 & 14.6$\pm$2.53 &  \cmark \\
        PIP Torque to backdrive (Nm) & 0.1 & 0.022$\pm$0.005 & \cmark \\
        PIP Maximum Speed (rad/s) & 4.5 & 21.6$\pm$0.2 & \cmark \\
        MCP-X Peak Torque (Nm) & 0.86 & 1.87$\pm$0.07  & \cmark \\
        MCP-X Bandwidth(Hz)@0.86Nm & 10.1 & 25.2$\pm$0.1 &  \cmark \\
        MCP Torque to backdrive (Nm) & 0.1 & 0.094$\pm$0.006 & \cmark \\
        MCP Maximum Speed (rad/s) & 4.5 & 9.8$\pm$0.3 & \cmark \\
        Width (Finger) (mm) & 22.4 & 15 & \cmark \\
        Width (Knuckle) (mm) & 22.4 & 36 & \xmark \\
        Height (Finger) (mm) & 22.4 & 9 & \cmark \\
        Height (Knuckle) (mm) & 22.4 & 28 & \xmark \\
        Length of Finger (mm) & 104.2 & 90.5 & \cmark \\
    \end{tabularx}
    \label{tab:desiderata}
\end{table}

\section{Proposed Finger Design}
\label{sec:proposed-design}
We now describe an actuator and finger design that closely matches the desired functional characteristics obtained in the previous section. 


\subsection{Motor}
\label{sec:motor-design}
The maximum torque generated by an actuator, $\tau_{motor}$, can be estimated by using the gear ratio $N$, and the expected electromagnetic shear stress, $\tau_{em}$, that can pass through the area of a motor's rotor-stator interface, which has a diameter $D$ and length $L$. The value of $\tau_{em}$ can be obtained from engineering tables \cite{Fundamentalsofdesign}.  

\begin{equation}
\tau_{motor}=N\tau_{em}\frac{D}{2}\left(\pi DL\right)
\label{eqn:motor_tau}
\end{equation}

To ensure compliance in our finger design, we aim to minimize the torque required to backdrive the motor by keeping the gear ratio low  \cite{bhatia2019direct}.
The PIP joint is limited to a width and height of 22.4mm each. Allocating space for a pulley, bearing, and structural elements leaves room for a motor with rotor-stator interface of 20mm diameter (D) and 10mm in length (L). For the higher-end stock motors that are not actively cooled, a reasonable value for $\tau_{em}$ is 0.2 bar \cite{Fundamentalsofdesign}. With these specs, the motor is expected to produce 0.15 Nm of torque according to Eq.\ref{eqn:motor_tau}. Our ideal motor then necessitates a 6:1 gear ratio for the PIP joint, and 7:1 for the MCP joint to match our torque requirements. 

However, the actual motors available did not perfectly match these ideal specifications. For the PIP joint, we selected a motor with an outer diameter of 28mm and a length of 14mm, slightly larger than the ideal motor size. This motor necessitated an 8:1 gear ratio to match our torque requirements. For the MCP joint, we chose a larger motor with an outer diameter of 34.5mm and a length of 15.7mm. We used a 13.6:1 gear ratio for this motor, despite exceeding our torque requirements by nearly 2x, because it allowed us to experiment with higher torques. We didn't do this for the PIP joint because it was difficult to fit a larger gear ratio in the same space. The next iteration of the MCP joint will reduce the gear ratio back to only what is necessary. Further details on the motors are available in the Appendix.

\subsection{Transmission Design}
\label{subsec:transmission-stiffness-design}

\begin{figure}[t!]
\centering

\vspace{0.3cm}
\includegraphics[width=6cm]{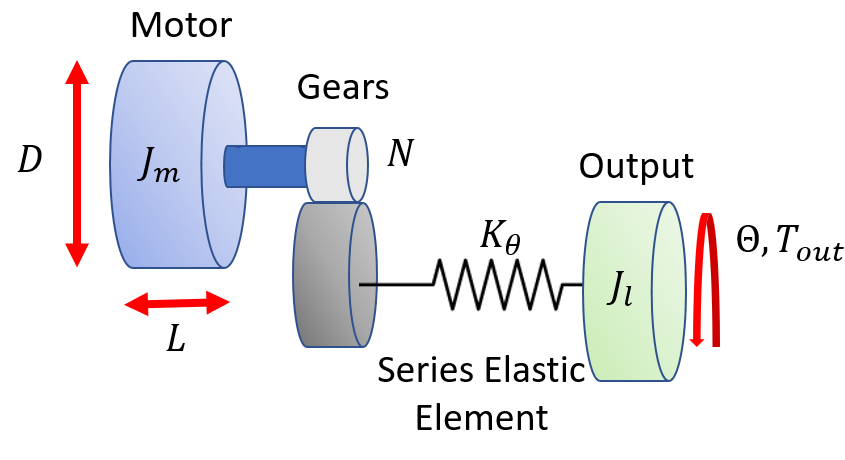}
\caption{Classic model for a series elastic actuator.}
\label{fig:sea-illustration}
\end{figure}

We designed our transmission to be compact yet strong enough to transmit necessary torques and handle impacts, aiming for a 1M cycle lifespan, which should last 2 years based on 12 hours a day usage and 60 pick and place maneuvers per hour, enough for many applications such as home robots. This transmission utilizes the full length of the fingers, featuring three reduction stages. Three stages were needed to keep the gears small, since larger reductions in each stage require larger gears. The first stage employs a belt to increase the center-to-center distance between the motor and subsequent transmission stages, giving them more room. These later stages use gears to maximize the load capacity within the available volume. Gear strength is estimated using the Lewis factor equation (eqn.\ref{eqn:lewis_Equation}) \cite{lewis_spurgear}, taking into account gear diameter $D_{gear}$, width $w_{\text{gear}}$, module $m$, safety factor $SF$, material yield strength $\sigma_y$, the Lewis factor $\gamma$ (from engineering charts), and the gear ratio to the output $N_{gear}$.
 
\begin{equation}
 T_{\text{strength}} = \frac{D_{gear}}{2SF}m \gamma \sigma_y w_{gear} N_{gear}
 \label{eqn:lewis_Equation}
\end{equation}

We opted for 0.5 module gears, since they are widely available and have adequate strength. A 30 tooth, 0.5 module gear of 15mm diameter made of carbon steel ($\gamma$ = 0.35, $\sigma_y$ = 415MPa), according to eqn.\ref{eqn:lewis_Equation}, would have a maximum driving torque of 0.54 Nm for every 1mm of width in the gear using a safety factor of 2. To satisfy the torques we require at the output stage, we then expect to need gears of 2mm in width. While a better estimate of the safety factor required for 1M cycles can be obtained through further analysis of manufacturing and operating conditions, 2 was sufficient for determining the approximate size gears necessary. Then, the actual expected strength for at least 1M cycles was obtained from the seller. Each gear in the transmission was analyzed similar to above, varying the materials and sizes to satisfy the torques they will undergo.

\subsection{Transmission Stiffness}
The stiffness of our transmission determines the bandwidth and the torque experienced during impact. The transmission can be made very stiff to increase bandwidth, but it also results in higher torques during impact, necessitating a stronger, thus larger, transmission. On the other hand, to make the finger resistant to impact, the transmission can be made less stiff, but then it will result in low bandwidth, which is insufficient for agile behaviors. We desire to achieve the maximum possible bandwidth while limiting impact torque to be lower than the strength of our transmission.
The impact torque can be calculated by analyzing a full-speed collision of the finger against an immovable object \cite{katz2019mini}. In this case, all the inertial energy from the motor ($ \frac{1}{2}J_mN^2\dot{\theta^2}_{max}$)  will be absorbed by transmission with stiffness $k_\theta$ which lead to an angular deflection in the finger by the amount $\Delta\theta_{max}$. Here $J_m$ is motor inertia, $N$ is the gear ratio, $\dot{\theta}_{\max}$ is the maximum speed of the motor. The maximum motor displacement is therefore:    
\begin{align}
{U}_{spring} =\frac{1}{2}k_{\theta}\Delta\theta_{max}^2 \ &= \frac{1}{2}\left(J_m\right)N^2\dot{\theta}_{max}^2\\     
\Rightarrow \Delta\theta_{max} &= N\dot{\theta}_{max}\sqrt{\frac{J_m}{k_\theta}}
\end{align}

The peak torque from the collision, $\tau_{collision}$ will be experienced at this maximum spring displacement and can be calculated as following: 
\begin{equation}
\tau_{collision}=\ k_{\theta}\Delta\theta_{max} =N\dot{\theta}_{max}\sqrt{k_{\theta}J_m} 
\end{equation}

We can equate the maximum torque experienced to the strength of the transmission, $\tau_{collision} = \tau_{strength}$, in our case, the torque at which the gears break which can be used to estimate maximum $k_{\theta}$.

\begin{equation}
k_{\theta}^{\max} = \frac{\tau^2_{strength}}{ (N\dot{\theta}_{max})^2J_m} 
\end{equation}

The minimum stiffness possible for our system is when we match our minimum bandwidth requirement. The bandwidth for our system can be found by approximating it as the natural frequency \cite{katz2019mini}, $\omega_n$, which can be calculated with the motor inertia $J_m$, gear ratio $N$, and stiffness $k_{\theta}^{min}$, as follows:

\begin{align}
B \approx \omega_{n}&=\ \sqrt{\frac{k_{\theta}^{min}}{{J}_mN^2}}\\
k_{\theta}^{\min}&=B^2J_mN^2
\end{align}

Using the values of $B, J_m, N, \tau_{strength}, \dot{\theta}_{max} $ from our parts and specifications, we obtain a range of possible values for the stiffness between: $k_{\theta}^{min}$ = 2.24 Nm/rad and $k_{\theta}^{max}$ = 10.16 Nm/rad. This range is low enough that it requires a spring in series to reach this regime of stiffness. While there has been success with propioceptive actuators without a spring in series for quadrupeds, for our robotic hand, a spring was required to reach the low stiffness levels necessary to sustain impacts.

\subsection{Design of the Elastic Element}
\label{sec:sea-design}
\begin{figure}[h]
\centering
\includegraphics[width=\linewidth]{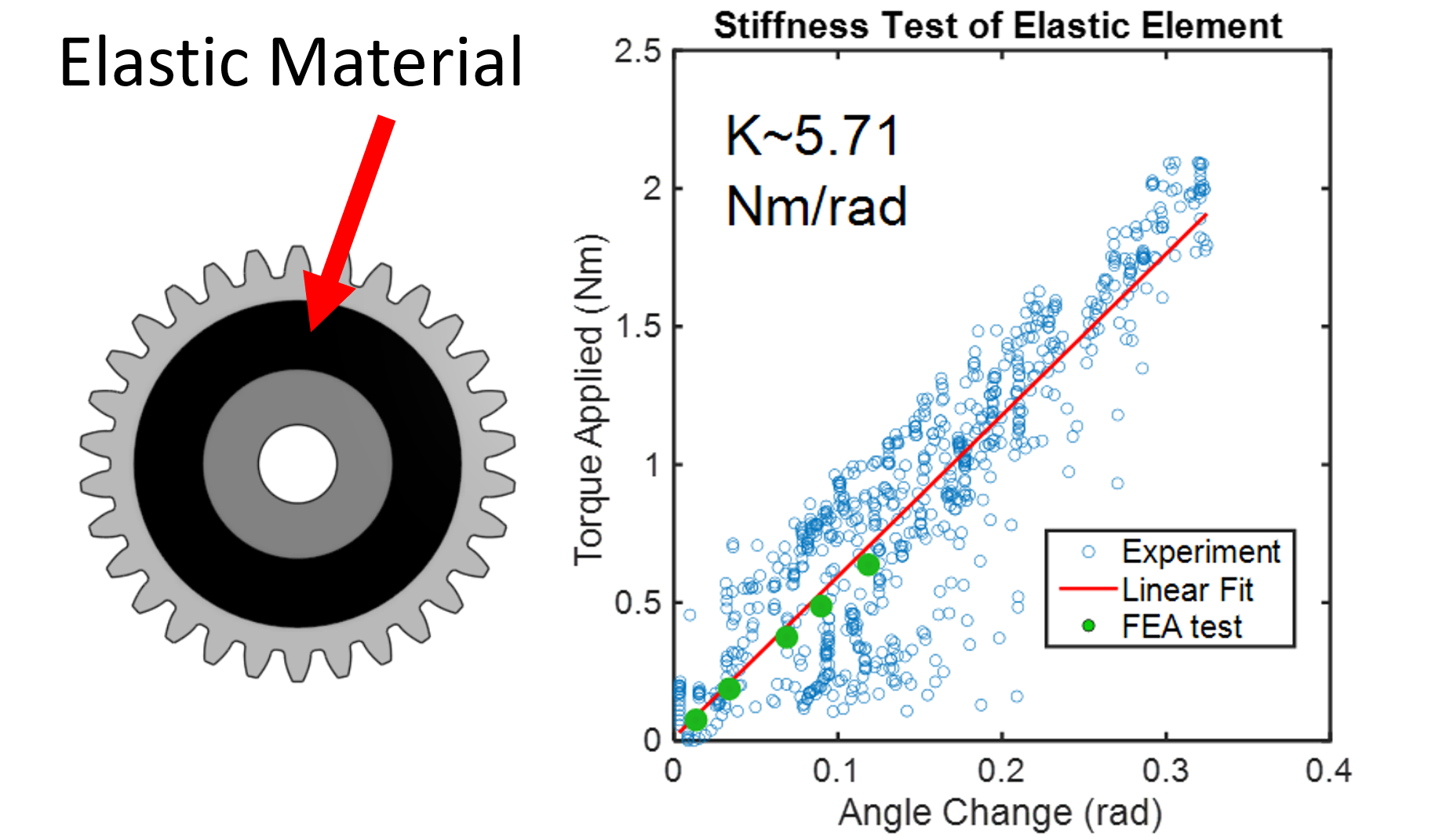}
\caption{Left: A gear with elastic material in the annulus between the ID and the OD helps compliance without increasing the overall size of the finger. Right: Experimental data characterizing the elastic element stiffness shows that it produces a linear response.}
\label{fig:sea}
\end{figure}

The main challenge in inserting a spring in the transmission is packaging so it satisfies the size requirements. We obtain this by customizing one gear in the transmission to have the spring element be embedded inside it as shown and described in Figure~\ref{fig:sea}. 

We tested the linearity of our elastic element by running both finite element analysis (FEA) experiments and collecting real-world data. We measured the displacement of the elastic element with applied torques using a high resolution camera. The results, shown in Fig. \ref{fig:sea}, show that the spring stiffness is linear.

\subsection{4-bar linkage}
\begin{figure}[t!]
\centering
\vspace{0.2cm}
\includegraphics[width=\linewidth]{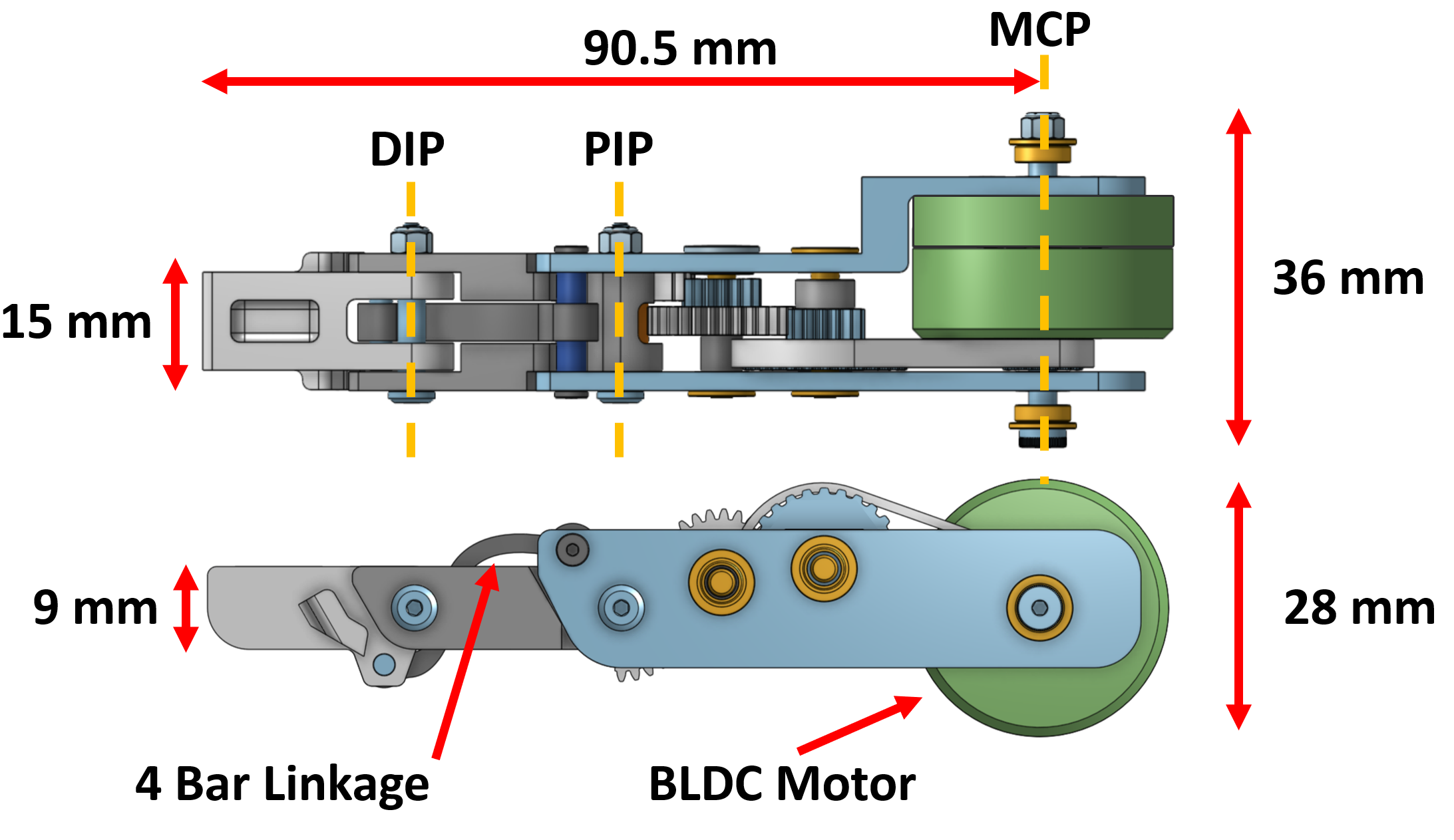}
\caption{Everyday robotic finger measurements and design, without MCP actuator.}
\label{fig:finger-design}
\end{figure}

Inspired by the Psyonic Ability hand, we incorporated a 4-bar linkage between the DIP and PIP joints, as shown in Figure ~\ref{fig:finger-design}. This choice makes the DIP joint rotate when the PIP joint rotates, mimicking the natural curling motion of human fingers.

\section{Experiments}
Our experiments have two primary goals: first, to characterize the performance of our actuator/finger along the five key design traits. The second goal is to qualitatively evaluate the performance of the robotic fingers on a few representative tasks that require compact size, force application, speed, and compliance. We chose these tasks as picking and placing dishes in a dish rack, picking strawberries, and picking a napkin lying flat on a table. 

\begin{figure*}[t!]
    \centering
    \vspace{0.2cm}
    \includegraphics[width=\textwidth]{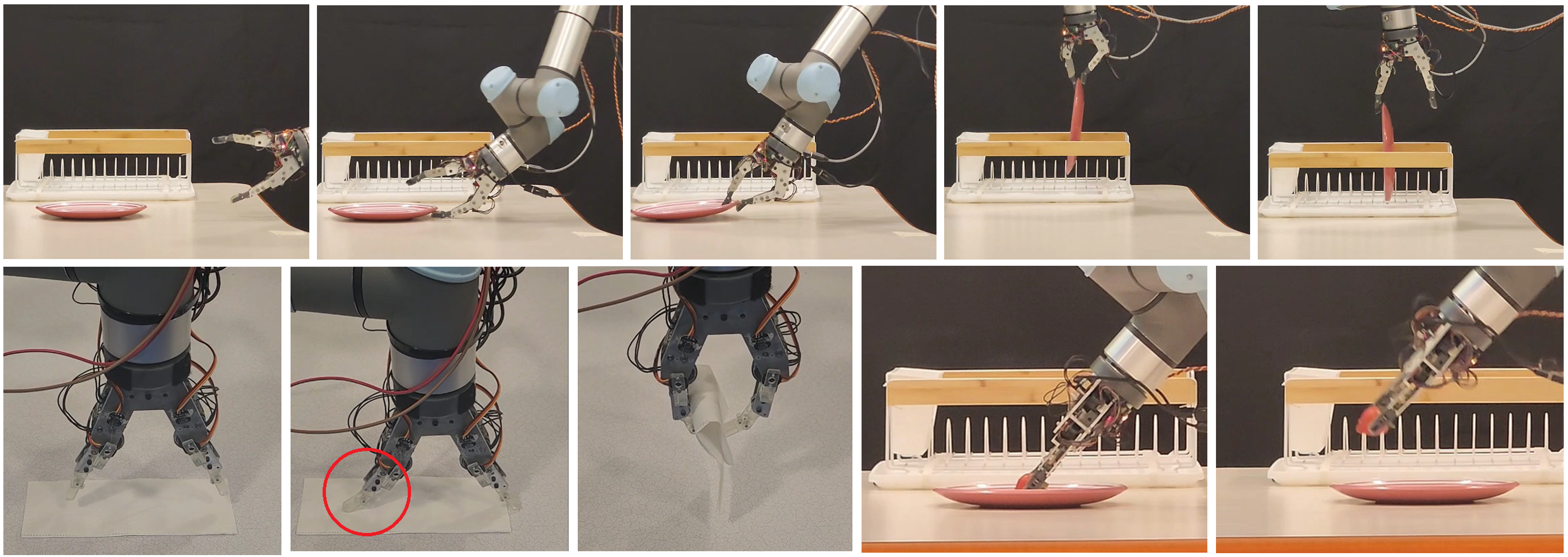}
    \caption{Qualitative tests run with the Everyday two-finger hand: (1) picking up and placing a dish. (2) picking up a flat napkin, and (3) picking up a strawberry without damaging it. }
    \label{fig:tasks-visualization}
\end{figure*}

\subsection{Characterization experiments}
For the characterization experiments, we mounted one finger on a Universal Robot's UR5 arm.
\begin{figure}[b!]
\centering
\includegraphics[ width=\linewidth]{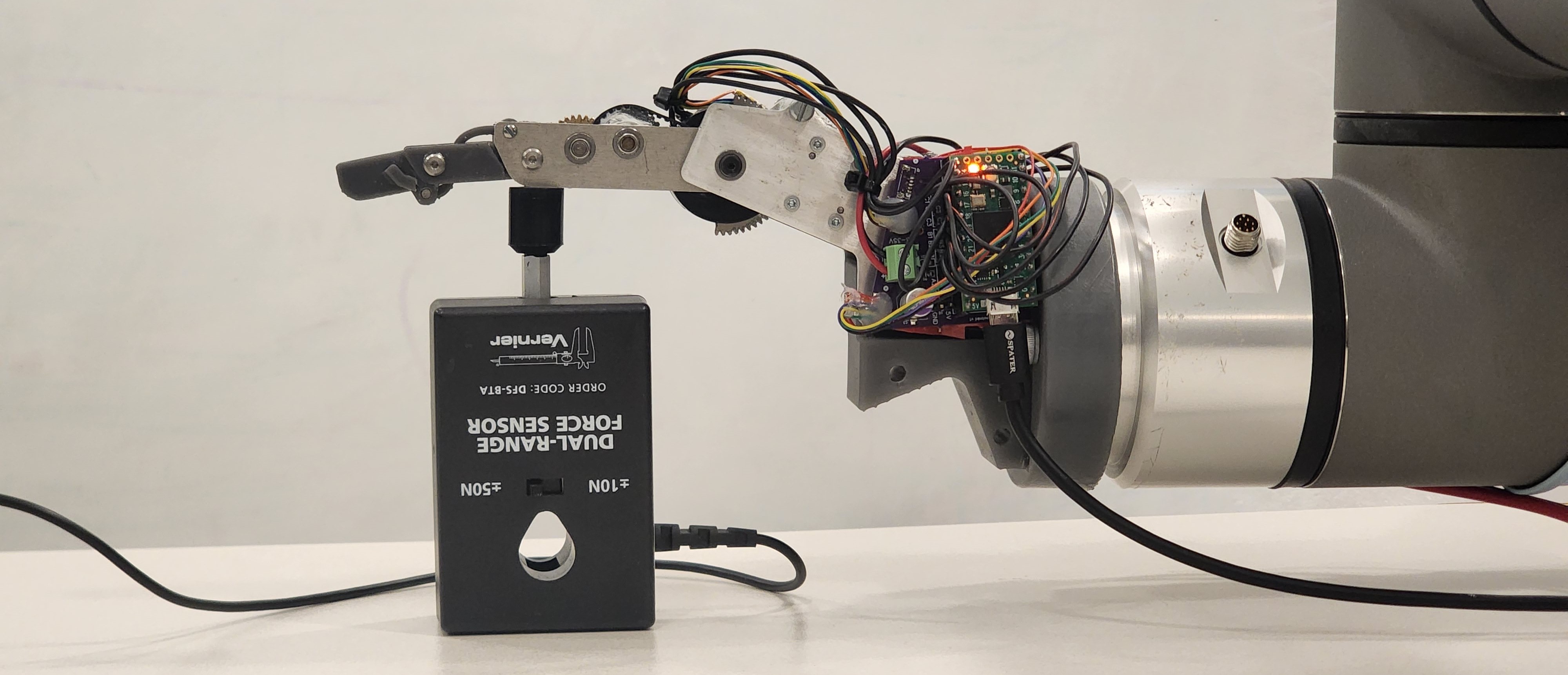}
\caption{Test setup for sensing maximum force and bandwidth.}
\label{fig:bandwidth-eval}
\end{figure}

\subsubsection{Torque and bandwidth characterization}

We placed a Vernier Dual-Range Force sensor 45 mm away from the MCP-X joint to measure the maximum force it could exert. From this, we can calculate the maximum torque. Simultaneously, we measure the rise time of the torque output, calculated as the time it takes to go from 10\% to 90\% of the desired torque, which can be used to estimate the torque bandwidth of the system at a given torque. Given the rise time,  $t_r$, the bandwidth is commonly approximated as~\cite{816121abandwidth_risetime}:  $B (Hz) = 0.35/t_r (s)$. We used the same method for the PIP joint.

Based on our data, the MCP joint has a peak torque of $1.87\pm0.07$Nm , and the PIP Joint of $0.68\pm0.01$Nm. The bandwidth changes for each commanded torque, and various datapoints can be seen on Figure ~\ref{fig:Optimization}, which clearly show we can meet the bandwidth requirements of our sample tasks.

\subsubsection{Compliance}
To measure backdrivability, we used the same setup as the experiments above. We lowered UR5 arm with the mounted finger onto the force sensor at $\sim$ 1m/s, the maximum speed at which the UR5 can move, starting from 5cm above the sensor, with the motor off until the finger moved about 45 degrees. After five trials, we measured the torque to push back the finger at 0.094$\pm$0.006 Nm for the MCP joint, and 0.022±0.005 Nm for the PIP joint.

\subsubsection{Speed}
The maximum speed of the PIP and MCP joints was measured by moving the finger from one end of its limit to the other. The finger position is measured using the internal magnetic encoders. We found the maximum speed to be 9.8$\pm$0.3 rad/s for the MCP Joint and 21.6$\pm$0.2 rad/s for the PIP Joint. 

\subsubsection{Compactness}
Our finger was more compact than the functional requirements for most of the finger. However, due to the size of the motor, the knuckle was larger than the desired width.

\subsection{Qualitative Experiments}
\label{sec:qualitative-results}
For the qualitative experiments, we mounted two fingers on a UR5 robotic arm, and hard-coded a fixed trajectory for each task. 

\subsubsection{Fast dish Pick and place}
\label{subsec:dish-pickup}
This task highlights unique advantages of the hand: it's compactness and compliance. While many robotic hands that are large have to rely on long routines to pick up dishes by pushing them against a wall \cite{zhou2022learning} \cite{florence2019selfsupervised} or towards the edge of the table, our fingers comply against the table after making contact and slide underneath the dish lip to grasp. Additionally, fingers must apply requisite forces to hold the dish tight while it is transported to the dish rack. Compliance again becomes critical while placing the dish, allowing the dish to settle into one of the dish rack slots without being overconstrained by the fingers. Visualization of this task is provided in Figure~\ref{fig:tasks-visualization}.

\subsubsection{Paper Napkin Pickup}
\label{subsec:napkin-pickup}
Picking flat objects with no clearance from the surface necessitates compliance to increase the contact surface between the robotic finger and the object being picked. In theory, a napkin can be picked by commanding a non-compliant gripper, but would require sub-millimeter precision to retain contact with the napkin as it lifts it up (a feat that is challenging for visual perception). 
Because our design is compliant, we purposefully drive the manipulator into the table, which bends the fingers against the table increasing the contact surface area, and then we close the fingers while moving up. Figure~\ref{fig:tasks-visualization} shows that our hand can successfully pick paper napkins. 

\subsubsection{Strawberry Pickup}
\label{subsec:strawberry}
We chose this task to demonstrate the speed and compliance of the fingers simultaneously. 
We lowered the gain of the position controller and along with the compliance of our fingers, we successfully grasped the strawberry over a dozen times without any visible damage to the strawberry. 

\section{Discussion and Future Work }
\label{sec:discussion}

To keep the finger phalanges compact and still meet the performance required, our robotic finger is larger than the average human hand at the knuckle and palm. Additionally, our finger only has 2 DOF. With a 3rd DOF, the size of the palm would increase. It would be difficult to fit 5 fingers onto a hand in a future iteration, but a 3 and maybe 4 finger hand is still possible. To make a full 5 finger hand, we may have to wait for more torque-dense actuators to become available.

While the main components of the hand (bearings, gears, and motor) have been chosen to last 1M cycles, the remainder of the parts and overall assembly have not been as thoroughly analyzed or tested for longevity. Future work involves making a 3-finger version that is fully dexterous, a wrist, and integrating tactile sensing. While our method for estimating the force and bandwidth requirements is a step in understanding performance requirements, further work needs to be done to gather data from humans directly at the joint level for more tasks. Capturing at the joint-level would reduce any potential source of error from mapping FT data to joint torques, and a larger dataset would be a great tool for designing hands that go beyond everyday tasks.

\section{Acknowledgements }
\label{sec:Acknowledgements}
We thank the members of the Improbable AI lab for the helpful discussions and feedback on the paper. This research was supported by funding from Toyota Research Institute and Hyundai Motor Company. We acknowledge support from ONR MURI under grant number N00014-22-1-2740.

\section*{Author Contributions}
\textbf{Rubén Castro Ornelas} led the project, came up with the high-level ideas, and developed/fabricated the design

\textbf{Tomás Cantú} Coded the optimization for figuring out joint torque trajectories, made custom motor controller PCBs 

\textbf{Isabel Sperandio} performed modeling and FEA on the structural components of the hand to ensure durability.

\textbf{Alexander H. Slocum} advised the modeling of the first prototype and elastic element within his class (MIT's 2.77)

\textbf{Pulkit Agrawal} advised the project, facilitated technical discussions throughout, and revised the paper.

 \section{Appendix}

\subsection{Optimization tasks}

\begin{table}[H]
\centering
\begin{tabular}{|c|c|c|c|}
\hline
\textbf{Name} & \textbf{Size} & \textbf{Palm} \\
\hline
stir with spatula & Large & L-Pinch & \cmark \\
\hline
sprinkle, shake pepper & Medium & Tripod1 &\cmark \\
\hline
spread/oil & Small & M-Pinch &\xmark \\
\hline
vertical cut & Large & L-Pinch &\cmark \\
\hline
use spoon to pick up & Small & Tripod3 & \xmark \\
\hline
pizza wheel & Medium & Tripod2 &\cmark \\
\hline
use black brush & Medium & M-Pinch &\cmark \\
\hline
spear object using fork & Small & L-Pinch & \cmark \\
\hline
stir water using spoon & Small & M-Pinch &\cmark \\
\hline
fasten screw with screwdriver & Medium & M-Pinch &\cmark \\
\hline
loosen screw with screwdriver & Medium & M-Pinch & \cmark \\
\hline
unlock lock with key & Small & Tripod1 & \xmark \\
\hline
fasten nut with wrench & Medium & L-Pinch &\cmark \\
\hline
use paint brush to dip and spread & Medium & M-Pinch  & \cmark \\
\hline
use hammer to hammer in nail & Large & L-Pinch & \cmark \\
\hline
brush teeth & Medium & M-Pinch &\cmark \\
\hline
use file to file wooden thing & Medium & L-Pinch & \cmark \\
\hline
comb hair & Medium & L-Pinch &\cmark \\
\hline
scrape substance from surface & Large & L-Pinch & \cmark \\
\hline
peel cucumber/potato & Medium & L-Pinch &  \cmark \\
\hline
slice cucumber & Medium & L-Pinch &\cmark \\
\hline
flip bread & Medium & Tripod3 & \xmark \\
\hline
use spoon to scoop and pour & Medium & M-Pinch & \cmark \\
\hline
shave object & Medium & L-Pinch & \cmark \\
\hline
use roller to roll out dough & Large & M-Pinch & \cmark \\
\hline
loosen nut with wrench & Medium & L-Pinch &\cmark \\
\hline
scoop and pour with measuring spoon/cup & Medium & M-Pinch & \cmark \\
\hline
insert peg into pegboard & Small & Tripod1 & \xmark \\
\hline
brush powder accross grey tray & Small & M-Pinch & \cmark \\
\hline
insert straw through to-go cup lid & Small & M-Pinch &\cmark \\
\hline
\end{tabular}
\caption{Size of handle used (Small: , Medium, or Large: ) as well as presence of palm for torque trajectory optimization analysis. The task data comes from \cite{huang2018dataset}.}
\label{tab:grip_description}
\end{table}

\subsection{Grasps}

\begin{figure}[h!]
\centering

\vspace{0.3cm}
\includegraphics[width=8cm]{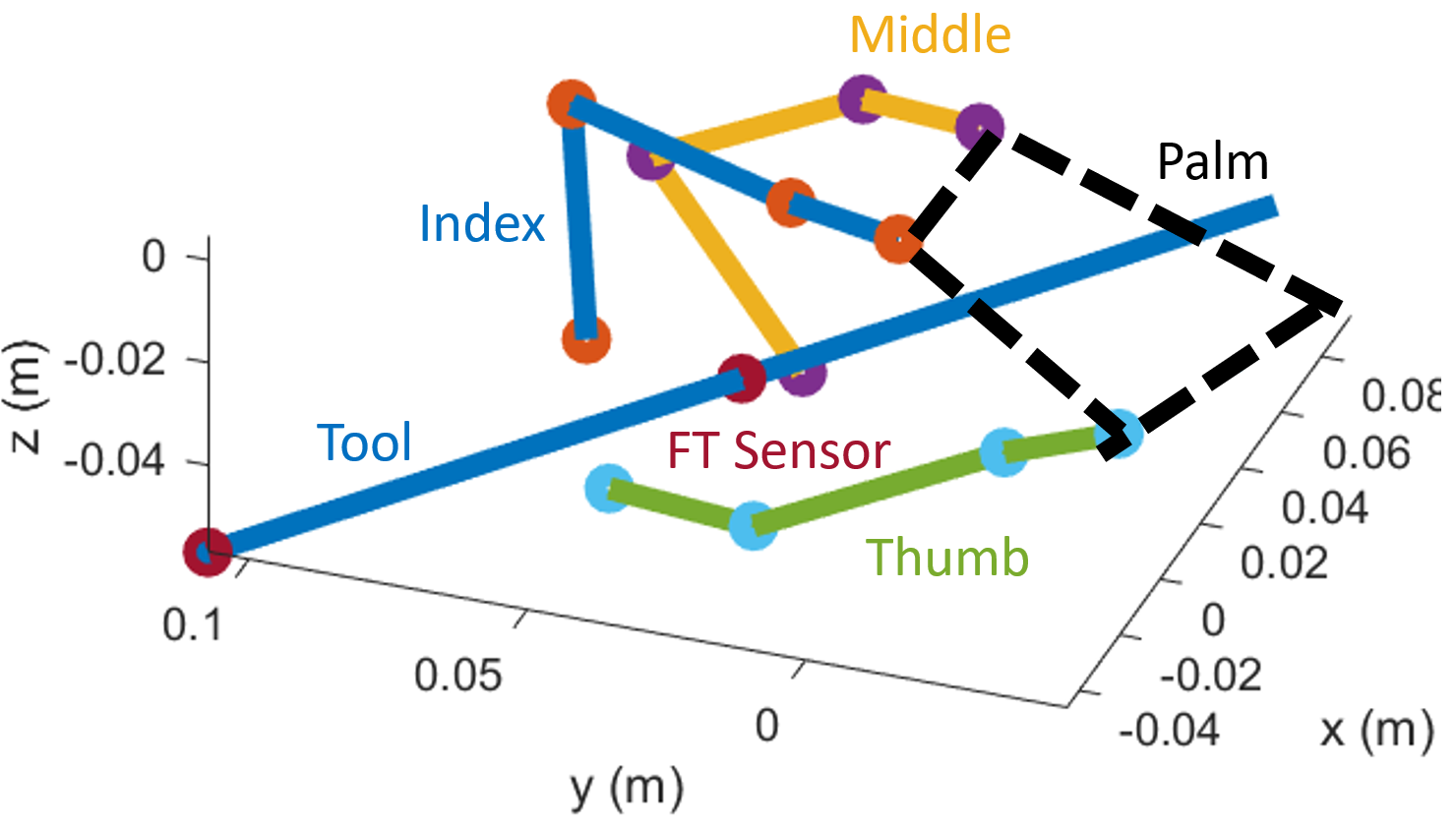}
\caption{Medium Pinch grasp}
\label{fig:medium_pinch}
\end{figure}

\begin{figure}[h!]
\centering

\vspace{0.3cm}
\includegraphics[width=8cm]{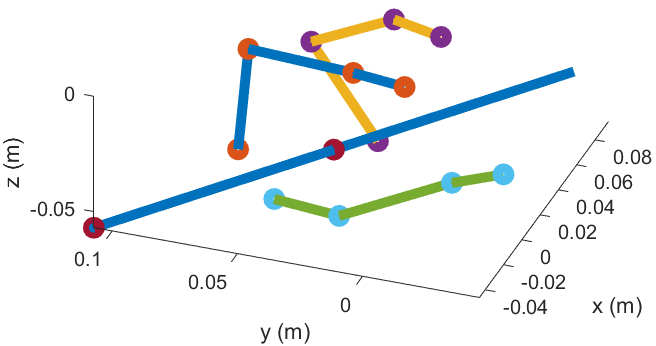}
\caption{Large Pinch grasp}
\label{fig:large_pinch}
\end{figure}

\begin{figure}[h!]
\centering

\vspace{0.3cm}
\includegraphics[width=8cm]{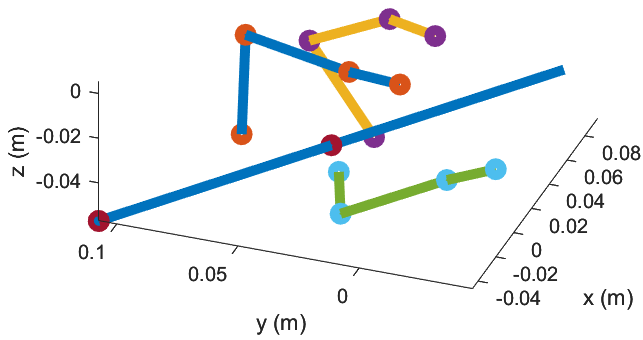}
\caption{Tripod grasp}
\label{fig:Tripod_Grasp}
\end{figure}

\begin{figure}[h!]
\centering

\vspace{0.3cm}
\includegraphics[width=8cm]{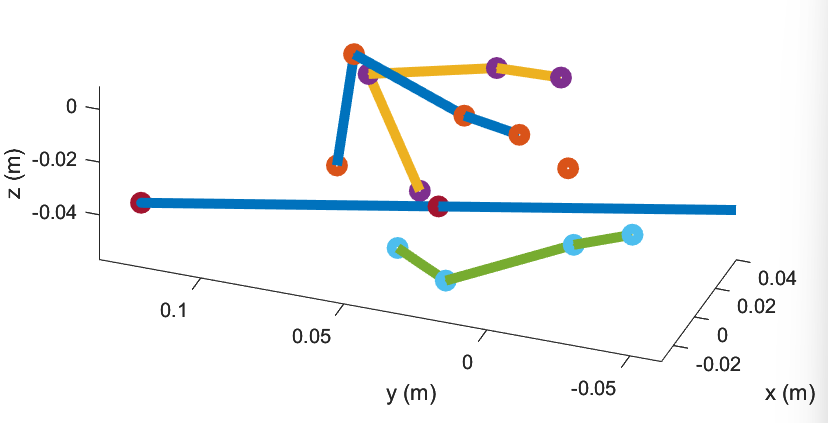}
\caption{Tripod2 Grasp}
\label{fig:Tripod_Grasp2}
\end{figure}

\begin{figure}[h!]
\centering

\vspace{0.3cm}
\includegraphics[width=8cm]{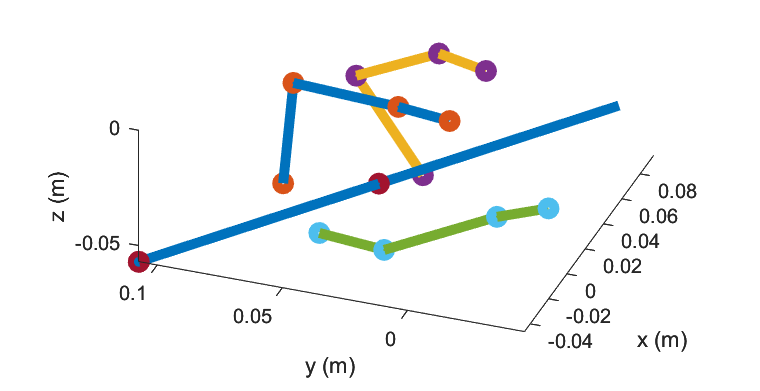}
\caption{Tripod3 Grasp}
\label{fig:Tripod_Grasp3}
\end{figure}

\subsection{Motors}

We use two different brushless DC motors. 

The MCP joint motor is Cubemars' GL30 gimbal-type motor. This motor was larger than what we needed for this joint, but was the closest to matching our desired specifications. The provided specifications can be seen below in Table \ref{tab:gl30_kv290_specs}.

\begin{table}[H]
\centering
\caption{Provided GL30 Motor Specifications}
\begin{tabular}{|l|l|}
\hline
\textbf{Specification} & \textbf{Value} \\
\hline
Motor Weight & 41g \\
\hline
Rated Voltage & 12V \\
\hline
Rated Current & 2.3A \\
\hline
Rated Torque & 0.08Nm \\
\hline
Rated Speed & 2200rpm \\
\hline
Overall Diameter & 34.5mm \\
\hline
Length & 15.7mm \\
\hline
Stall/Peak Torque & 0.28Nm \\
\hline
\end{tabular}
\label{tab:gl30_kv290_specs}
\end{table}

The PIP joint motor is a 2204 Gimbal-type motor.The stated specifications are below in Table \ref{tab:technical_data_2204}. We found these specifications to be off. The motor could not output as much power as we hoped, but was still enough to meet the requirements for the PIP joint.

\begin{table}[H]
\centering
\caption{Provided 2204 Gimbal Motor Specifications}
\begin{tabular}{|l|l|}
\hline
\textbf{Specification} & \textbf{Value} \\
\hline
Rated Voltage & 7.4V \\
\hline
Maximum continuous current & 1.3A \\
\hline
Kv & 260 RPM/V \\
\hline
Maximum power & 15W \\
\hline
Overall diameter & 28mm \\
\hline
Length & 15mm \\
\hline

\end{tabular}
\label{tab:technical_data_2204}
\end{table}


\addtolength{\textheight}{-12cm}   





\printbibliography 

\end{document}